\documentclass[review,12pt]{elsarticle}

 \pdfoutput=1
\usepackage{amssymb}
\usepackage{amsmath}
\usepackage{url}
\usepackage{float}
\usepackage{hyperref}
\usepackage{multirow}
\hypersetup{
	colorlinks=true,
	linkcolor=cyan,
	filecolor=blue,      
	urlcolor=red,
	citecolor=green,
}
\usepackage{tabularx,booktabs}
\usepackage{bm}

\usepackage{lineno}
\usepackage{graphicx}
\usepackage{subfigure}
\usepackage{float}
\usepackage[ruled,linesnumbered]{algorithm2e}
\journal{Pattern Recognition}

\begin{document}
\begin{frontmatter}



\title{Line Graph Contrastive Learning for Link Prediction}

\author[label_1]{Zehua Zhang\corref{cor1}}
\ead{zhangzehua@tyut.edu.cn}
\cortext[cor1]{Co-first authors and Corresponding author}
\author[label_1]{Shilin Sun\corref{cor2}}
\cortext[cor2]{Co-first authors}
\ead{shilin_sun01@163.com}
\author[label_2]{Guixiang Ma}
\ead{guixiang.ma@intel.com}
\author[label_3]{Caiming Zhong}
\ead{zhongcaiming@nbu.edu.cn}
\address[label_1]{College of Information and Computer,Taiyuan University of Technology,
            Yuci District, 
            Jinzhong,
            030600, 
            Shanxi,
            China}
            \address[label_2]{
          Intel Labs, Hillsboro, Oregon, 97229,USA}
\address[label_3]{College of Science and Technology, Ning bo University,
           Cixi, 
            Ningbo,
            315300, 
            Zhejiang,
            China}

\begin{abstract}
Link prediction tasks focus on predicting possible future connections. Most existing researches measure the likelihood of links by different similarity scores on node pairs and predict links between nodes. However, the similarity-based approaches have some challenges in information loss on nodes and generalization ability on similarity indexes. To address the above issues, we propose a Line Graph Contrastive Learning(LGCL) method to obtain rich information with multiple perspectives. LGCL obtains a subgraph view by $h$-hop subgraph sampling with target node pairs. After transforming the sampled subgraph into a line graph, the link prediction task is converted into a node classification task, which graph convolution progress can learn edge embeddings from graphs more effectively. Then we design a novel cross-scale contrastive learning framework on the line graph and the subgraph to maximize the mutual information of them, so that fuses the structure and feature information. The experimental results demonstrate that the proposed LGCL outperforms the state-of-the-art methods and has better performance on generalization and robustness.
\end{abstract}


\begin{highlights}
\item We design a novel contrastive learning framework based on line graph to be  suitable for link prediction on sparse and dense graphs.
\item We propose a cross-scale contrastive learning strategy to maximize the mutual information between subgraph and line graph.
\item The dual perspectives contrastive progress to some extent avoids the problem of inconsistent prediction on the similarity based methods with single view.
\item Our comprehensive experiments on six datasets from diverse areas demonstrate that our model has better performance on generalization and robustness than the SOTA methods.
\end{highlights}


\end{frontmatter}


\section{Introduction}
\label{Introduction}
 Link prediction task is based on the topological definition of the network to predict the existence of links between nodes. It has been applied to various fields, such as product recommendations~\cite{anand2022integrating}, biological molecule interaction prediction~\cite{kishan2021predicting}, traffic forecasting~\cite{ZHANG2019308}, etc.
 
The current research on link prediction usually follows a kind of human intuition that the more similar the attributes or topological structure of two nodes are, the more likely that they have interactions with each other. Based on the common similarity principle, several network similarity methods have been proposed for link prediction task~\cite{kumar2020link}. And they are designed by minimizing the pointwise mutual information (PMI) of co-occurring nodes in random walk~\cite{AGIBETOV2023108977}. Besides, the idea of multi-level analysis is introduced to deal with graph structure data from the local and global levels. TOME~\cite{ZHONG20152699} proposes two refinement processes to obtain local information of cluster structure and incorporates some global information into the matrix with path-based transformation. Thereby, network similarity methods can also be classified from multiple perspectives of the local and the global. The Node Clustering Coefficient~\cite{WU20161} evaluates the clustering coefficients of all common neighbors of the target node pair and sums them to obtain the final similarity score of the node pair. Such methods can effectively handle link prediction task in dynamic networks, such as traffic networks. Whereas, extracting only local information will limit the ability to capture global similarities between nodes. In contrast, other studies utilize global topological information of a network to score the similarity of nodes, such as Katz~\cite{Katz195339}, Random Walk with Restart (RWR) ~\cite{4053087}, and Rooted Pagerank~\cite{BRIN1998107}. Except on sparsely unbalanced networks, the global structure based methods have shown better performance than based on local information. Furthermore, the global methods are not suitable for large-scale networks, specially with dense connections, due to huge computational costs. In addition, the major challenge on these similarity based methods does not avoid similarity measurement selection  and generalization limitation of a single index.

Obviously, another idea to improve the prediction accuracy  can be derived from edge information on graph to mine deeply the data to obtain richer information. With the development of deep learning on graph data,  researchers pay more attention to graph representation learning methods with the ability of learning graph topological information and enhancing node features~\cite{WANG2022108215}. Specially, the feature learning process for the node is based on the assumption that nodes with similar embedding representations will display similar structures. For example, HOGCN~\cite{kishan2021predicting} adopts different distance features on neighbors and shows excellent robustness in sparse interaction networks. Wang et al.~\cite{wang2022sparse} introduce the HAS method via Heterogeneous graph data Augmentation and node Similarity to solve sparse imbalanced link prediction. What's more, it is noteworthy that the representation and structure dual similarity assumption is not universal. For instance, some proteins with similar characteristics but may have a lower probability of connections~\cite{kovacs2019network}. To address such issues, SEAL~\cite{zhang2018link} converts the link prediction task into a graph classification task by extracting the subgraphs around the target links. Only the node information pooling is used to predict links in SEAL, the loss of node information will  bring disturbance to the accuracy on prediction. Therefore, reducing the information loss has become another challenge for current graph neural network-based methods. 

In comparison, LGLP~\cite{cai2021line} transforms prediction task into a node classification task, by combining line graphs with graph neural networks for link prediction to improve information transfer efficiency. LGLP can alleviate the problem of inefficient learning with sparse data, whereas the added edges during information transfer may also bring in noise. So it will limit the model's performance to ignore the balance of different levels of information.

Everything has two sides and the line graph transform is not an exception. On dense graphs, the noise generated by the excessive edges of line graph can adversely impact the prediction results~\cite{ZHONG201993}. By contrast, the information generated by the increased edges of the transformation can improve the prediction accuracy. So our motivation comes from how to make up for the lack of line graph conversion and improve the performance by self-supervised learning without additional information.  

It is hard or expensive to acquire data labels in many practical applications, whereas contrastive learning is an excellent self-supervised learning method to improve model performance with less labels. Current studies focus on designing diverse graph augmentation strategies to yield various representations of the same node. Via contrastive learning loss, the contrastive learning methods maximize the consistency of anchor nodes with positive samples, and minimize the similarity of anchor nodes to negative samples~\cite{you2020graph}. Nonetheless, the methods mainly alter the graph structures or features instead of the type of graph to generate new views, and the whole progress of methods performs on the same task from single perspective.

To sum up, the paper introduces the Line Graph Contrastive Learning (LGCL) method for link prediction task to  be compatible with sparse and dense graph mining problems. By converting a subgraph into a line graph, the edges in the subgraph can be converted to nodes in the line graph. If edges have shared nodes, there are edges between the corresponding line graph nodes. LGCL obtains the edge embedding information directly from a graph encoder. Meanwhile, the link prediction task is directly converted into a node classification task. Aiming at the information redundancy of line graphs, based on the
principle of mutual information maximization, we take subgraphs and line graph nodes as two different representations of the links, and propose a novel subgraph-line graph node contrastive paradigm to balance them. Our contributions are summarized as follows:
\begin{itemize}
    \item We design a novel contrastive learning framework based on line graph. The LGCL method has less information loss than the traditional methods, both suitable for link prediction on sparse and dense graphs. 
\item We propose a cross-scale contrastive learning strategy to maximize the mutual information between subgraph and line graph. The dual perspectives contrastive progress to some extent avoids the problem of inconsistent prediction on the similarity based methods with single view.
\item We have conducted a comprehensive comparison and analysis with mainstream benchmark methods on six public datasets. Besides, the ablation and parametric sensitivity analysis confirm the effectiveness of our algorithm. The experimental results show that the proposed LGCL has better performance on generalization and robustness than the SOTA methods.
\end{itemize}
\section{Related Work}
\begin{figure}[H]
\centering
\includegraphics[width=5.5 cm]{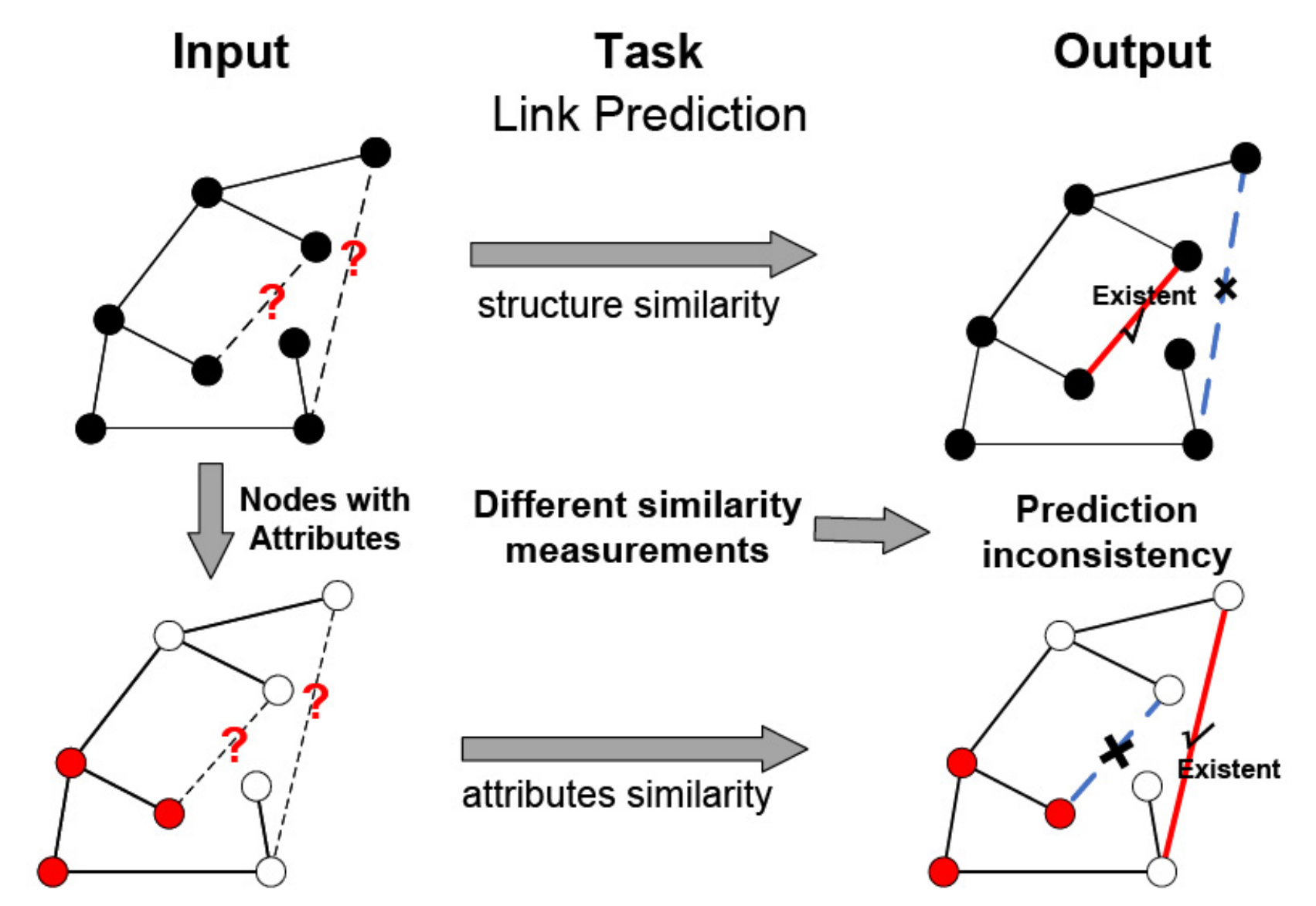}
\caption{Link Prediction\label{fig_link}}
\vspace{-1.0em}
\end{figure}

The main purpose of link prediction is to predict the existence of missing links, as shown in Fig.\ref{fig_link}. Many similarity measurements have been proposed in recent studies~\cite{kumar2020link}, and the similarity based on topology structure is easy to use but relatively simple. So prediction accuracy based on structural similarity is often closely related to similarity index selection and network features. In addition, the node attributes of topology network that can usually be obtained directly may be hidden or even incomplete. Therefore, link prediction based on fusion attributes is prone to conflict between structure and attribute similarity, resulting in inconsistent prediction. With the development of deep learning, researchers further expand link prediction methods by combining network embedding and graph representation learning.

\subsection{Network Similarity-based Methods}%
Network similarity methods predict the existence of links according to the predicted node pair's topology structure and node attribute, a link with a high score indicates that it is more likely to exist~\cite{kumar2020link}. The methods of network similarity can be divided into two main categories, local and global methods. Local network similarity methods like Adamic-Adar~\cite{adamic2003friends} score how similar common attributes of two nodes are to predict links between nodes. Similarity Regularized Nonnegative Matrix Factorization (SRNMF)~\cite{wang2017similarity} considers local features and network geometric information by combining similarity based structure and latent features for link prediction in binary networks. Global network similarity methods consider the whole network like SimRank~\cite{jeh2002simrank}, which take a random walk at two separate nodes and score them depending on the time the walker meets each other. And L3~\cite{kovacs2019network} evaluates all node pairs based on the number of paths of length 3 between them. Others, including Katz~\cite{Katz195339} and Rooted Pagerank~\cite{BRIN1998107}, utilize global network information to score node pairs. In general, similarity-based methods assume that nodes prefer to make associations with nodes that have less variance in their surrounding structural layouts, which limites their performance. 
\subsection{Network Embedding-based Methods}
To reduce noisy or redundant information which can be defined as the curse of dimensionality, network embedding methods first aim to learn dense and continuous representations of nodes in low-dimensional spaces, and retain the inherent structural information. They represent the nodes as low-dimensional vectors, so that the data will be dimensioned to take advantage of the effective information. Skip-gram~\cite{mikolov2013distributed} predicts the words in its window except for the central word when the target word is given. DeepWalk~\cite{perozzi2014deepwalk} integrates Skip-gram, an approach for word vectors in natural language processing, into the network, so it can utilize truncated random walks to gain structural information about nodes and learn potential embedding representations of nodes in the network. Node2vec~\cite{grover2016node2vec} adopts random walk, but it is biased as a consequence of the trade-off between DFS and BFS. Du et al.~\cite{du2020cross} propose a novel cross-network embedding model that extends the skip-gram, it alternately performs link prediction and network alignment through joint optimization. Such methods, on the other hand, are restricted to learning topological information about the network structure and are unable to effectively aggregate information about network nodes.
\subsection{Graph Representation Learning-based Methods}
Graph representation learning aims to use the attribute characteristics of nodes and the structural features of graphs to learn the representation of nodes and the distribution of links in a graph~\cite{HU2021107745}. For instance, GCN~\cite{welling2016semi} learns node representation mainly by aggregating first-order neighbor information. SEAL~\cite{zhang2018link} converts the link prediction task into a graph classification task by extracting the subgraphs of the links. However, all of these approaches rely on node embedding information, though they perform well enough in node representation learning, they can not obtain edge information directly for link prediction. 
\subsection{Line Graph-based Methods}
With the development of graph neural networks, researchers are beginning to explore the structural properties of line graphs and apply them to various tasks. CensNet~\cite{jiang2019censnet} adopts line graph to obtain edge embedding information, and it proposes two new convolution operations to embed edge and node information into the same latent feature space concurrently. It performs well in semi-supervised node classification, multi-task graph classification, and graph regression. DHCN~\cite{xia2021self} models session-based data with hypergraphs, then it converts hypergraphs into line graphs, and it uses contrastive learning as auxiliary tasks to maximize the mutual information represented by the two views to improve the recommendation task. LGLP~\cite{cai2021line} proposes to convert link subgraphs into line graphs, but 
the edges of line graphs may generate noise when the training set is large, and their over-reliance on single-view information affects the performance of the model. Since graph convolutional networks usually rely only on node features, GAIN~\cite{GHARAEE2021108174} incorporates highly representative edge features into graph convolutional networks to classify road types through line graph transformation. 

To sum up, there is no such research to maximize the mutual information between line graphs and subgraphs by contrastive learning to predict links.
\subsection{Graph Contrastive Learning-based Methods}
Graph contrastive learning (GCL) evaluates the similarity of samples, reduces the distance between similar samples, and increases the distance between different samples. Its core idea is to maximize the MI between similar graph instances and minimize the MI between different instances~\cite{Liu_2022}. Existing works can be grouped into two kinds of contrastive learning methods: same-scale, and cross-scale.
\begin{enumerate}
    \item[a.]\textbf{Same-Scale Contrast}: Same-Scale Contrast can be categorized as Graph-Graph Contrast and Node-Node Contrast. GraphCL~\cite{you2020graph} uses four types of data augmentation to perturb the graph for an augmented graph, then it learns the graph representation with a shared encoder, and finally maximizes the mutual information of the two graphs. Unlike Graph-Graph Contrast as above, GRACE~\cite{GRACE} processes augmented graphs through a shared encoder to obtain node representations of two views, and maximizes the mutual information between node representations of the two views. GCA~\cite{Zhu_2021} follows the framework of GRACE, it proposes adaptive augmentation strategies.
    \item[b.]\textbf{Cross-Scale Contrast}: Cross-Scale Contrast can also be described as node-graph contrast, so it has significant advantages in assisting the graph encoder in obtaining local and global graph information. DGI~\cite{velickovic2019deep} first feeds the original graph and the augmented graph into two different graph encoders, thereby obtaining graph representation and node representation. MVGRL~\cite{hassani2020contrastive} follows DGI and generates two views using graph diffusion and subgraph sampling techniques. 
    Then it maximizes the MI between node representations and graph representations across scales.
\end{enumerate}

By analyzing the link prediction task from different task perspectives, we propose a cross-scale contrastive method of subgraph-line graph node contrast. Different from graph-node contrast, our model uses two kinds of graphs to represent the links with subgraphs and line graph nodes respectively. And to maximize the MI between them, it obtains global and local information of links by different encoders.
\section{Method}
\subsection{Problem Formulation}
Given an interaction network $\textbf{\emph{G}}(\textbf{\emph{V}},\textbf{\emph{E}},\textbf{\emph{X}},\textbf{\emph{A}})$, where $\textbf{\emph{V}}$ represents the set of nodes that correspond to entities, and $\textbf{\emph{E}}\subseteq{\textbf{\emph{V}}\times\textbf{\emph{V}}}$ denotes the set of edges which indicates the existence of interaction between two entities in $\textbf{\emph{V}}$. $\textbf{\emph{X}}$ is a feature matrix where each node in $\textbf{\emph{V}}$ is encoded as a predefined attribute vector. The adjacency matrix of $\textbf{\emph{G}}$ as $\textbf{\emph{A}}$ represents interaction relations of node pairs. For an arbitrary node pair  $v_i,v_j$, the model aims at predicting the relationship between them and training the mapping function $f(\cdot)$ as shown in Eq. (\ref{eq}).
\begin{equation}
f({v_i},{v_j}) = \left\{ \begin{array}{l}
1,{e_{{v_i},{v_j}}} \in \textbf{\textit{E}} \\
0, Otherwise 
\end{array} \right. \label{eq}
\end{equation}
where $e_{{v_i},{v_j}}$denotes the edge connected by node $v_i$ and node $v_j$, and  $f({v_i},{v_j})$ equals $1$ if there exists a link from $v_i$ to $v_j$, and $f({v_i},{v_j})$ equals $0$ otherwise.            

Most of contrastive learning methods derive from mutual information (MI), which is aimed to measure the interdependence of diverse variables, and they aim to maximize MI. Through contrastive learning as formulated in Eq. (\ref{eq_score}), the model maximizes MI between line graph and subgraph information. The line graph transformed from a subgraph of an arbitrary predicted node pair is the positive sample, line graphs converted from other subgraphs are negative samples.
\begin{equation}
max\quad\mathcal{I}
(f_{S}(\bm{{X}_{S}}),f_{L}(\bm{{X}_{L}}))\label{eq_score}
\end{equation}
where $\bm{X_{S}}$ and $\bm{X_{L}}$ denote the subgraph feature matrix, line graph feature matrix separately. Correspondingly, ${f}_{S}$ and $\emph{f}_{L}$ correspond to the encoders of subgraph and line graph, $\mathcal{I}(\cdot)$ is an estimator for mutual information. In this paper, we use NCE objective to alternatively maximize $\mathcal{I}(\cdot)$’s lower
bound. Then the total loss $\mathcal{L}_{total}$ is calculated by weighing the subgraph supervised loss $\mathcal{L}_{S}$, line graph supervised loss $\mathcal{L}_{L}$, and contrastive learning loss $\mathcal{L}_{CON}$ as shown in Eq. (\ref{eq2}).
\begin{equation}
    \mathcal{L}_{total} = f_{agg}(\mathcal{L}_{S},\mathcal{L}_{L},\mathcal{L}_{CON})\label{eq2}
\end{equation}
where $f_{agg}$ is an aggregation function, which multiplies various losses by different coefficients and sum them.
\subsection{Overview}
\begin{figure}[H]
\centering
\includegraphics[width=13.5cm]{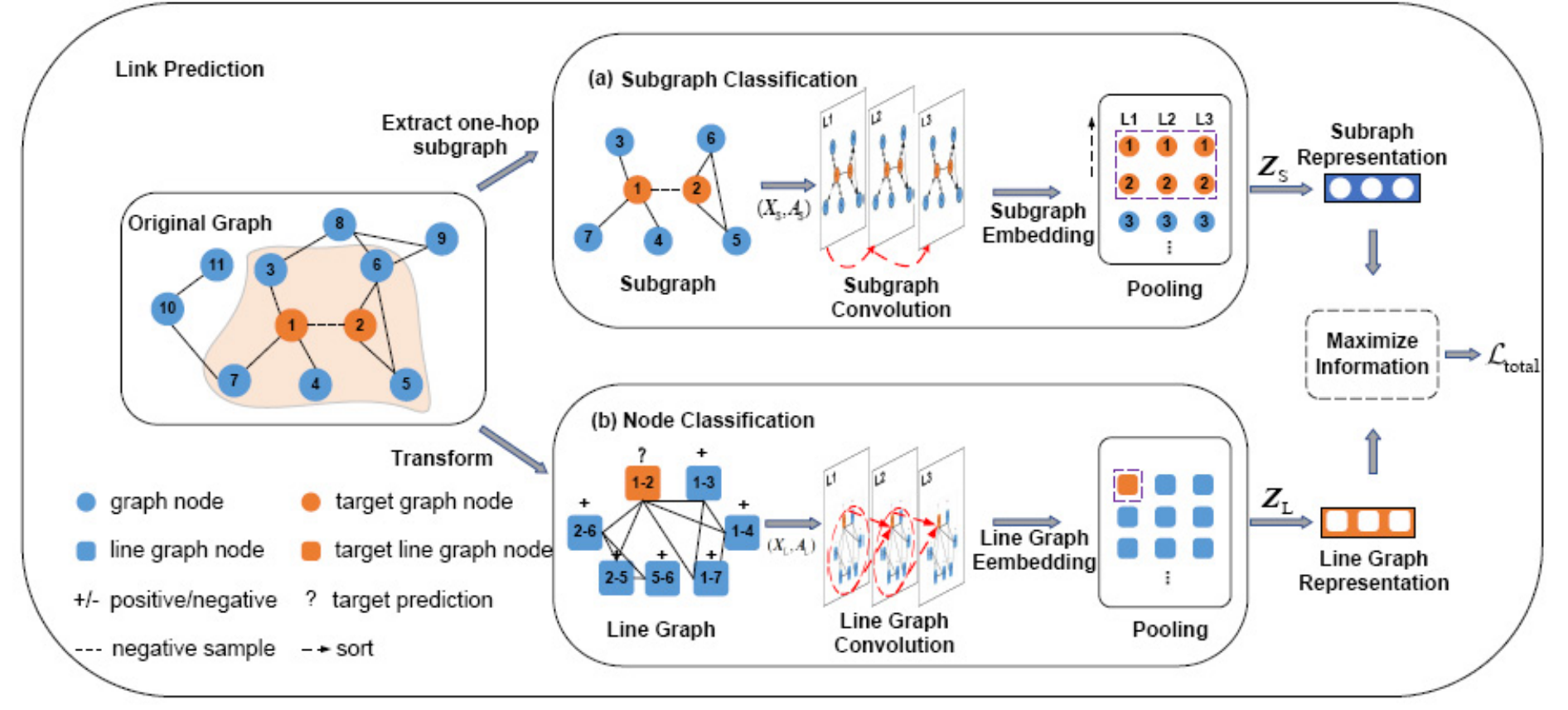}
\caption{Overview of the Line Graph Contrastive Learning framework.\label{fig1}}
\end{figure}

Fig.\ref{fig1} illustrates the overall framework of LGCL. The line graph transformation component samples the original graph to generate subgraphs centered on target predicted links and transforms them to line graphs. Next, the encoders encode the subgraph and line graph to obtain different kinds of representations for common links. Eventually, the contrastive learning component balances the information between two views.

\subsection{Line Graph Transformation}
LGCL adopts subgraph information for link prediction, and balances subgraph and line graph information by contrastive learning.

\emph{Subgraph Extraction}: The $h$-hop subgraph of the target prediction link is extracted from the node pairs, and the subgraph information is acquired, with $h$
denoting the number of hops as Eq. (\ref{eq3}).
\begin{equation}
{\bm{{V}_{S}}} = \{ v|\min ({\mathop{\rm d}\nolimits} (v,{v_i}),{\mathop{\rm d}\nolimits} (v,{v_j}) \le h)\} ,\forall v \in {\textbf{\emph{V}}}\label{eq3}
\end{equation}
where $d(\cdot)$ represents the function which calculates  the shortest path length between two nodes, and the new set of $h$-hop subgraph nodes $\bm{V_{S}}$ is generated by centering on arbitrarily two nodes $v_{i},v_{j}$ in , and the feature matrix of the subgraph $\bm{X_S}$ could be derived by $\bm{V_S}$. Edge set can be obtained as Eq. (\ref{eq4}).
\begin{equation}
    \bm{E_{S}} = \{ e_{{v_i}{\rm{,}}{v_j}}\}\cap \bm{E} ,\forall {v_i},{v_j} \in \bm{{V}_{S}}\label{eq4}
\end{equation}
where $\bm{E_{S}}$ denotes the set of edges, and the adjacency matrix of the subgraph $\bm{A_{S}}$ can be obtained from $\bm{E_{S}}$, and $e_{v_{i},v_{j}}$ denotes the edge between $v_{i}$ and  $v_{j}$. Finally, the $h$-hop subgraph centered on the predicted link $v_{i},v_{j}$ is obtained. 

\emph{Line Graph Transformation}: To transform a subgraph to a line graph, first we transform the subgraph's edges to the line graph's set of nodes, as illustrated in Eq. (\ref{eq5}). We form the line graph node feature by concatenating the two node features connected by the edge, which has the least information loss compared with the mean and weighted sum. If any two nodes of a line graph corresponds to two edges of a subgraph share a common node, then the line graph's nodes form an edge~\cite{cai2021line}. $\textbf{\emph{L}}$ is the set of line graph edges, and $\bm{N_{L}}$ is the set of line graph nodes, and $\bm{X_{L}}$ represents the feature matrix of line graph nodes as shown in  Eq. (\ref{eq5}). 
\begin{equation*}
   \bm{{N}_{L}}=\{e_{v_{i},v_{j}}\},\forall{e_{v_{i},v_{j}}}\in{{\bm{{E}_{S}}}}
\end{equation*}
\begin{equation*}
    \bm{X_{L}} =  {\mathop{\rm concate}\nolimits} ({x_i},{x_j})|\forall {e_{({v_i},{v_j})}} \in \bm{{N}_{L}}
\end{equation*}
\begin{equation}
    \bm{L} = \{l_{e_{({v_i},{v_j})},e_{({v_i},{v_k})}}{\rm{|}}{\{{v_i},{v_j}\}} \cap {\{{v_i},{v_k}\}} \ne \emptyset \} ,\forall e_{({v_i},{v_j})},e_{({v_i},{v_k})} \in \bm{N_{L}}\label{eq5}
\end{equation}

The identification of the node in the line graph indicates the link in the subgraph, and the number of edges in the line graph increases exponentially compared to the number of edges in the original graph as shown in Eq. (\ref{eq6}). $ Num(\cdot)$ is the function to calculate the number of elements in the set. 
\begin{equation}
        Num(\bm{L}) = {\sum\nolimits_{i = 1}^{Num(\bm{V_{S}})}} \frac{1}{2}{Deg({v_i})} ^2 - {\mathop{Num}\nolimits} (\bm{E_{S}}),\forall v_{i} \in \bm{V_{S}}\label{eq6}
\end{equation}
 where $Deg(\cdot)$ denotes a function to calculate the degree of nodes. Compared with original graph, the node information in the line graph is more comprehensive, and the information transfer of the line graph in the graph neural network is more efficient, because the node information of each line graph is combined with the information of two nodes in the original subgraph.

 \emph{Time Complexity Analysis:} For the original graph with n edges, in order to form the adjacency matrix of the line graph, we first need to sample one edge, and then determine whether the remaining $n-1$ edges have a common node with it, so the time complexity of constructing the line graph is $O (n^2)$. The complexity of message-passing depends on the number of edges in the graph. For example, a star graph with $n$ edges has $n^2$ edges after being converted into a line graph, so the time complexity is $O (n^2)$~\cite{jo2021ehgnn}.
\subsection{Graph Encoder}
The graph encoder $f(\cdot)$ is separated into two parts: a subgraph encoder $f_{S}(\cdot)$ and a line graph encoder $f_{L}(\cdot)$, where the graph encoder handles the feature and adjacency matrices of the two views. The representation of subgraphs and line graphs is conducted independently as $\textbf{\emph{Z}}_{S},\textbf{\emph{Z}}_{L}$ in Eq. (\ref{eq7}).

\begin{equation}
\bm{{Z}_{S}} = {f_{S}}(\bm{{X}_{S}}, \bm{{A}_{S}}),\quad  \bm{{Z}_{L}} = {f_{L}}(\bm{{X}_{L}},\bm{{A}_{L}})\label{eq7}
\end{equation}

  The convolution formula of the subgraph encoder is shown in Eq. (\ref{eq8}).
  \begin{equation}
    \bm{Z_{S}} = {f_{S}}({\widetilde {\textbf{\emph{D}}}^{{\bf{ - 1}}}}\widetilde {\textbf{\emph{A}}}\bm{X_{S}W})\label{eq8}
\end{equation}
where $\widetilde {\bm{A}}{\bf{ = }}\bm{A_{S}}{ + \bm{I}}$ represents the adjacency matrix of a graph with a self-loop. $\widetilde{\bm{D_{ij}}} = {\sum\nolimits_j \widetilde {\bm{A_{ij}}}}$, $\bm{X_{S}}$ is the node feature matrix, and $\textbf{\emph{W}}$ is the trainable graph convolution parameter matrix.

The graph encoder aggregates the node information in the local neighborhood of the graph to extract local substructure information. Extraction of substructure features by stacking multiple graph convolution layers is shown in Eq. (\ref{eq9}).
\begin{equation}
  \bm{Z_{S}^{h+1}} = f_{S}({\widetilde {\bm{D}}^{{\bf{ - 1}}}}\widetilde {\bm{A}}  \bm{Z_{S}^{h}}  \bm{W_{S}^{h+1}}),\quad \bm{Z_{S}^{1:h}} : = [\bm{Z_{S}^{1}}  \ldots \bm{Z_{S}^{h}}]\label{eq9}
\end{equation}
where $h$ represents the number of layer and $\bm{Z_{S}^{1:h}}$ which is obtained by concatenating representations at different layers.
Finally, the obtained graph is pooled through the SortPooling layer~\cite{zhang2018end} to obtain a graph-level representation, and a supervised loss ${{\mathcal{L}}_{\rm{S}}}$ is obtained as shown in Eq. (\ref{eq10}).
\begin{equation}
{{\mathcal{L}}_{\rm{S}}} =  - \frac{1}{{|\bm{T}|}}\sum\limits_{{({v_{i,}}{v_j})} \in \bm{T}} {{p_{s}}\log {{\widehat p}_{{s}}} + (1 - {p_{s}})} \log (1 - {\widehat p_{s}})\label{eq10}
\end{equation}
where $|\bm{T}|$ is the total number of node pairs in the training dataset, ${\widehat p_s}$ is the model's prediction for subgraph $s$, which is the $h$-hop subgraph centered on $v_i$ and $v_j$, and ${p_{s}}$ is the true result for the interaction between $v_i$ and $v_j$.

Line graph encoder adopts GCN~\cite{welling2016semi} which is shown in Eq. (\ref{eq_gcn}).
\begin{equation}
    {x_i}^{h + 1} = \sigma (\sum\limits_{{v_j} \in {V_{\rm{S}}}} {\frac{1}{{{c_{ij}}}}} {x_j}^{(h)}{w^{(h)}} + {b^{(h)}})\label{eq_gcn}
\end{equation}
where $x_{j}^{(h)}$ denotes features of node $j$ at layer $h$, $\sigma(\cdot)$ denotes non-linear activation layer, and $c_{ij}$ represents normalization factor. $w^{(h)}$ is the weight of the $h$ layer, $c^{(h)}$ is the $h$ layer's intercept, and $v_{j}$ is one of the surrounding nodes of node $i$.

After transforming into the corresponding line graph:
\begin{equation}
    {x_{{u_{({v_i},{v_j})}}}}^{h + 1} = \sigma (\sum\limits_{{v_{(a,b)}} \in {{\bf{N}}_{\bf{u}}}} {\frac{1}{{{c_{uv}}}}} {x_v}^{(h)}{w^{(h)}} + {b^{(h)}})\label{eq12}
\end{equation}
where $u_{v_{i},v_{j}}$ denotes the node $u$ in the line graph is transformed from the $v_{i},v_{j}$ edge in the original graph and $N_{u}$ is the set of neighbors of node $u$ in the line graph.

Finally, the pooling layer selects the nodes, thereby analyzing the graph information to acquire $\bm{{Z}_{L}}$, and the supervised loss of the line graph is derived from the Eq. (\ref{eq13}).

\begin{equation}
    {{\mathcal{L}}_{\rm{L}}} =  - \frac{1}{|\bm{T}|}\sum\limits_{{(v_{i},v_{j})} \in T} p{_u}\log \widehat{{p}_{u}} + (1 - {p}_{u})\log (1 - \widehat{{p}_{u}})\label{eq13}
\end{equation}
where $\widehat{{p}_{u}}$ denotes the presence of the predicted link $e_{v_{i},v_{j}}$. ${p}_{u}$ is the actual situation in which the link exists.
\subsection{Contrastive Learning}
\begin{figure}[H]
\centering
\includegraphics[width=5.5cm]
{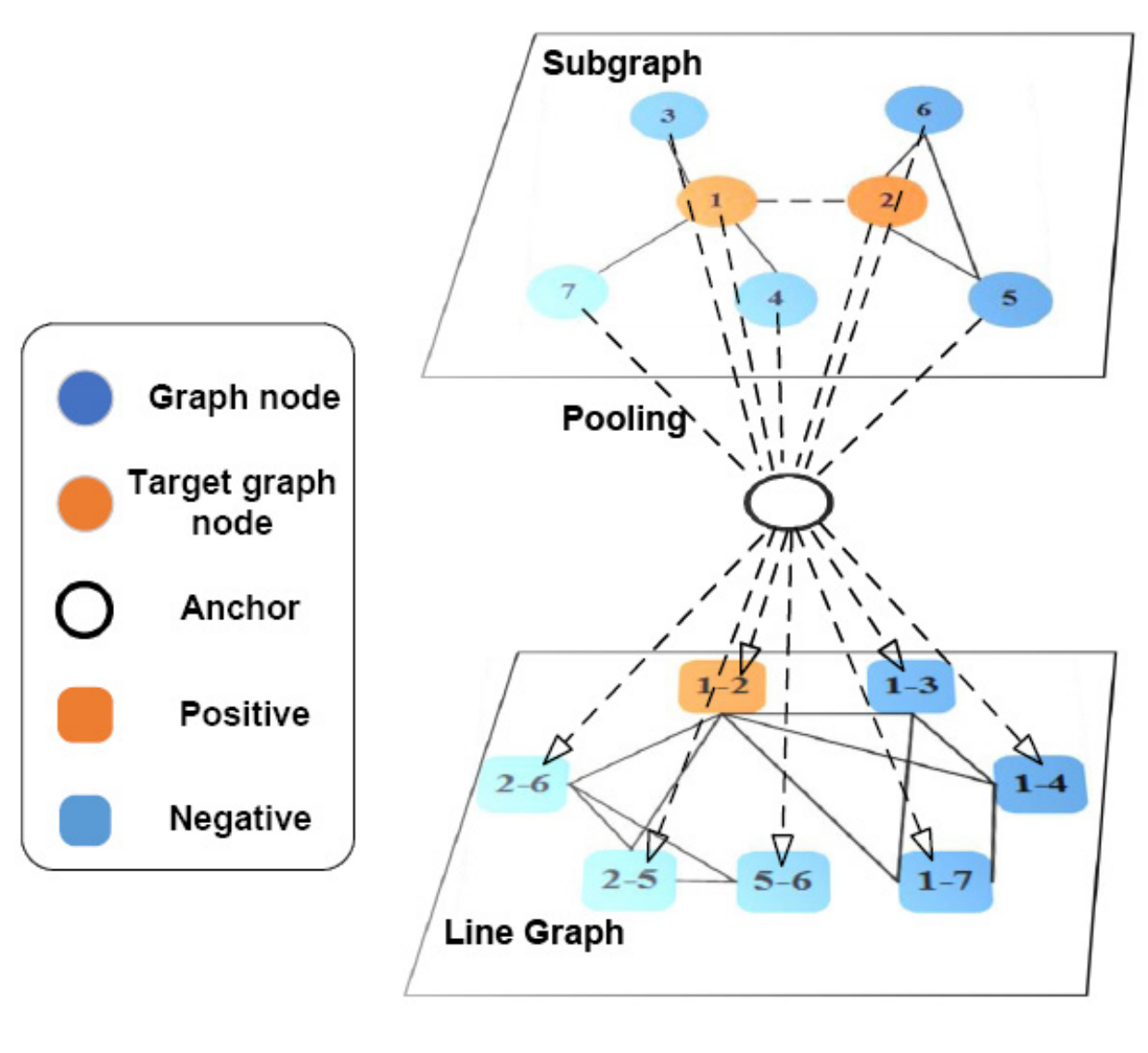}
\caption{The proposed diagram of line graph contrastive learning. Our contrastive learning consists of two views, the subgraph and the line graph views. As shown in the figure, the subgraph extracted from the link composed of nodes 1 and 2 is the anchor node, the line graph nodes composed of nodes 1 and 2 are positive samples, and the other nodes in the line graph are used as negative samples. \label{fig_CL}}
\end{figure} 
The number of edges in a line graph grows exponentially, while the noise increases as well. Contrastive learning mainly follows GraphCL~\cite{you2020graph}, which improves the model performance by  maximizing MI between two views. The generated line graph information is combined with the subgraph information to obtain the corresponding contrastive loss.

According to Fig.\ref{fig_CL}, take $z_{\rm{S}}^{(n)},z_{\rm{L}}^{(n)}$ in $\textbf{\emph{Z}}_{S},\textbf{\emph{Z}}_{L}$ respectively, to denote the two views of the $n$th graph in the small batch, the subgraph and line graph. Negative samples are generated from the other $N-1$ line graphs in the same batch. The cosine similarity function is denoted as ${\mathop{\rm sim}\nolimits} (z_{\rm{S}}^{(n)},z_{\rm{L}}^{(n)}) = \frac{{{{(z_{\rm{S}}^{(n)})}^{\rm{T}}}z_{\rm{L}}^{(n)}}}{{||z_{\rm{S}}^{(n)}||||z_{\rm{L}}^{(n)}||}}$, contrastive loss $\mathcal{L}_{CON}$ is shown in Eq. (\ref{eq_CL}).
\begin{equation}
    {{\mathcal{L}}_{{\rm{CON}}}} = \frac{1}{{{\rm{|}}\bm{T}{\rm{|}}}}\sum\nolimits_{n = 1}^{{\rm{|}}\bm{T}{\rm{|}}} { - \log \frac{{\exp ({\mathop{\rm sim}\nolimits} (z_{\rm{S}}^{(n)},z_{\rm{L}}^{(n)})/\tau )}}{{\sum\nolimits_{m = 1,m \ne n}^N {\exp ({\mathop{\rm sim}\nolimits} (z_{\rm{S}}^{(n)},z_{\rm{L}}^{(m)})/\tau )} }}}\label{eq_CL}
\end{equation}
where $|\bm{T}|$ denotes the number of node pairs in the training set and $\tau$ is a hyperparameter, and $N$ denotes the size of links in a training batch.

Finally, as shown in Eq. (\ref{eq_total}), the total loss function $\mathcal{L}_{total}$ is obtained by combining the self-supervised task loss with the supervised loss.

\begin{equation}
    {{\mathcal{L}}_{{\rm{total}}}} = \alpha {{\mathcal{L}}_{\rm{L}}} + (1-\alpha){{\mathcal{L}}_{\rm{S}}} + \beta {{\mathcal{L}}_{{\rm{CON}}}}\label{eq_total}
\end{equation}
where $\alpha, \beta$ are the hyperparameters used to balance the different losses. Our proposed algorithm is summarized as shown in Algorithm \ref{pseudocode}.

\begin{algorithm}[H]
\caption{Line graph contrastive learning algorithm}\label{pseudocode}
\SetKwInOut{Input}{input}\SetKwInOut{Output}{output}
\Input{Original graph  $\textbf{\emph{G}}(\textbf{\emph{V}},\textbf{\emph{E}},\textbf{\emph{X}},\textbf{\emph{A}})$,Training set $T$ }
\Output{Total loss $\mathcal{L}_{total}$}
\Repeat{All epochs have been trained}{
     \For{target node pair $({v_i},{v_j}) \in \bm{T}$}{
        Extract $h$-hop subgraph $\bm{G_{S}}(\bm{V_{S}},\bm{E_{S}},\bm{X_{S}} \bm{A_{S}})$ via Eq. (\ref{eq3})(\ref{eq4})\\
        Get line graph $\textbf{G}_{L}(\textbf{\emph{N}}_{L},\textbf{\emph{L}},\textbf{\emph{X}}_{L},\textbf{\emph{A}}_{L})$ via Eq. (\ref{eq5})\\
         Get subgraph representation $\textbf{\emph{Z}}_{S}$, line graph representation $\textbf{\emph{Z}}_{L}$ via Eq. (\ref{eq7})\\
         Calculate subgraph loss $\mathcal{L}_{S}$, line graph loss $\mathcal{L}_{L}$  via Eq. (\ref{eq10}) and Eq. (\ref{eq13})\\
          Calculate contrastive loss $\mathcal{L}_{CON}$ via Eq. (\ref{eq_CL})\\
          Calculate total loss $\mathcal{L}_{total}$ via Eq. (\ref{eq_total})\\
          Update model parameters\\
          
    }
}
Return: Total loss $\mathcal{L}_{total}$
\end{algorithm}

\section{Experiments}
\subsection{Datasets and Experiment Setup}
\begin{table}[h]
\begin{center}
\begin{minipage}{\textwidth}
\caption{Summary of datasets used in our experiments.}\label{dataset_tab}
\begin{tabular*}{\textwidth}{@{\extracolsep{\fill}}lcccccc@{\extracolsep{\fill}}}
\toprule%

	\textbf{Datasets}	& \textbf{Nodes}	& \textbf{Links}       & \textbf{Density} & \textbf{Area} \\
			\midrule
		Power~\cite{watts1998collective}   & 4941   & 6594  &0.054\%  &Power Network\\
HPD~\cite{peri2003development}  &8756    &32331   &0.084\%  &Biology\\
ADV~\cite{massa2009bowling} &5155   &39285   &0.296\%  &Social Network\\
Fdataset~\cite{gottlieb2011predict}  &906    &1933   &0.472\% &Biology\\
SMG~\cite{batagelj2006pajek}  &1024    &4916   &0.939\% &Co-authorship\\
ChCh-Miner~\cite{biosnapnets}  &1514    &48514   &4.236\% &Biology\\
\bottomrule
\end{tabular*}
\end{minipage}
\end{center}
\vspace{-2.0em}
\end{table}
We conduct experiments on six datasets, i.e., Power~\cite{watts1998collective}, HPD~\cite{peri2003development},  ADV~\cite{massa2009bowling}, Fdataset~\cite{gottlieb2011predict}, SMG~\cite{batagelj2006pajek}, ChCh-Miner~\cite{biosnapnets} from different areas to verify our proposed method's effectiveness. And these datasets have different density and scales as shown in Table~\ref{dataset_tab}. We perform all experiments on the Inspur heterogeneous cluster (GPU: 12*32G Tesla V100s, memory: 640G DDR2). And we deploy the LGCL framework with PyTorch, PyGCL$\footnote{\url{https://github.com/PyGCL/PyGCL}}$, and PyTorch Geometric(PyG). During the training process of the model, we adopt Optuna to optimize the parameters. To evaluate the effectiveness of link prediction, we adopt Area Under the Curve (AUC) and Average Precision (AP) as evaluation metrics. The datasets used for the experiments are available, and the code will be made public in the following work$\footnote{\url{https://github.com/ShilinSun/LGCL}}$.
The AUC value is equivalent to the expected probability of positive samples ranked ahead of uniformly chosen random negative samples, and AUC and AP are shown in Eq. (\ref{AUC}) and Eq. (\ref{AP}). 
\begin{equation*}
  FPR = \frac{FP}{FP+TN}
\end{equation*}
\begin{equation*}
  TPR = \frac{TP}{FP+FN}
\end{equation*}
\begin{equation}
  AUC = \sum_{k=1}^{n}FPR(k)\Delta TPR(k) \label{AUC}
  \end{equation}
where TP,TN,FP FN represent the number of true positive, true
negative, false positive and false negative samples in link prediction, respectively. False Positive Ratio (FPR) represents the proportion of false positive samples to the total number of negative samples, True Positive Ratio (TPR) denotes the proportion of true positive samples to the number of positive samples, and $n$ denotes the number of samples.
\begin{equation*}
  PPV = \frac{TP}{FP+TP}
\end{equation*}
\begin{equation}
  AP = \sum_{k=1}^{n}TPR(k)\Delta PPV(k) \label{AP}
  \end{equation}
where Positive Predictive Value (PPV) denotes the proportion of true positive samples to all those who have positive test results, and $n$ denotes the total number of samples.
\subsection{Baselines}
We compare our proposed approach with three types of benchmark methods as follows:
\begin{enumerate}
    \item Network Similarity-based Methods:
    \begin{itemize}
        \item Katz index~\cite{Katz195339} is based on the set of all paths, it sums directly over the set of paths, and decays exponentially by length to give more weight to the shorter paths.
        \item PageRank (PR)~\cite{BRIN1998107} calculates the similarity between two nodes, it iteratively computes the PageRank vector to obtain the similarity of a node with respect to other nodes.
        \item SimRank (SR)~\cite{jeh2002simrank} measures the similarity between any two nodes based on the topological information of the graph.
    \end{itemize}
    \item Network Embedding-based Methods:

    \begin{itemize}
        \item Node2vec (N2V)~\cite{grover2016node2vec} extends DeepWalk by running a biased random walk based on breadth or depth-first search, to capture local and global network structure.
    \end{itemize}
    \item Graph Representation Learning-based Methods:
    \begin{itemize}
        \item SEAL~\cite{zhang2018link} represents the links as subgraphs, and the existence of the links is indicated by the labels of the subgraphs.
        \item LGLP~\cite{cai2021line} transforms the link prediction task into a node classification task by introducing line graph theory.
    \end{itemize}
\end{enumerate}
\subsection{Results and Analysis}
\subsubsection{Comparison with Baselines}
Table \ref{AUC_benchmark} and Table \ref{AP_benchmark} show that the performance of LGCL method has improved performance compared with the other methods. N2V uses prior knowledge to control wandering, which has a better performance compared to similarity-based methods when the network data is sparse, such as Power dataset in our expermient, but it has a poor performance when the network data is dense and the data is complex, because it uses prior knowledge to bias walking. The graph representation learning methods directly use the graph embedding information, and the performance is better compared to other methods. Experimental results show that LGCL achieves the best performance by integrating line graph and subgraph information.
\begin{table}[H]
\begin{center}
\begin{minipage}{\textwidth}
\caption{Performance comparison with baselines on  training percentages (80\%) 
 (AUC)}\label{AUC_benchmark}
\begin{tabular*}{\textwidth}{@{\extracolsep{\fill}}lcccccc@{\extracolsep{\fill}}}
\toprule%

Model & Power & HPD & ADV    \\
\midrule
Katz~\cite{Katz195339} & 59.59($\pm$1.51)   & 85.47($\pm$0.35) &92.13($\pm$0.21)  \\
PR~\cite{BRIN1998107}  & 59.88($\pm$1.51)   & 87.19($\pm$0.34)  &92.78($\pm$0.18)  \\
SR~\cite{jeh2002simrank} & 70.18($\pm$0.75)   & 81.73($\pm$0.37)   &86.18($\pm$0.22)\\
N2V~\cite{grover2016node2vec} & 70.37($\pm$1.15)   &79.61($\pm$1.14) &77.70($\pm$0.83) \\
SEAL~\cite{zhang2018link}  & 81.37($\pm$0.93)   & 92.26($\pm$0.09)   &95.07($\pm$0.13)  \\
LGLP~\cite{cai2021line} & 82.17($\pm$0.57)   & 92.58($\pm$0.08) &95.40($\pm$0.10) \\
LGCL (Ours)  & \textbf{83.08}($\pm$0.86)   & \textbf{93.67}($\pm$0.15) &\textbf{96.65}($\pm$0.18) \\
\bottomrule
\end{tabular*}
\end{minipage}
\end{center}
\vspace{-4.0em}
\end{table}

\begin{table}[H]
\begin{center}
\begin{minipage}{\textwidth}
\resizebox{\linewidth}{!}{
\begin{tabular*}{\textwidth}{@{\extracolsep{\fill}}lcccccc@{\extracolsep{\fill}}}
\toprule%

Model & SMG & ChCh-Miner & Fdataset \\
\midrule
Katz~\cite{Katz195339} &86.09($\pm$1.06)   & 91.58($\pm$0.12) &92.24($\pm$0.23) \\
PR~\cite{BRIN1998107}  &89.13($\pm$0.90)   & 87.53($\pm$0.15)  &93.31($\pm$0.07)\\
SR~\cite{jeh2002simrank} &78.39($\pm$1.14)   & 81.50($\pm$0.17)   &86.18($\pm$0.22) \\
N2V~\cite{grover2016node2vec}  &78.30($\pm$1.22)   &79.23($\pm$1.12) &75.57($\pm$0.83)\\
SEALL~\cite{zhang2018link}  &91.53($\pm$0.46)   & 97.56($\pm$0.29)   &92.93($\pm$0.13)\\
LGLP~\cite{cai2021line} &92.53($\pm$0.29)   & 97.23($\pm$0.07) &94.95($\pm$0.24)\\
LGCL (Ours) &\textbf{93.64}($\pm$0.17)   & \textbf{97.67}($\pm$0.13) &\textbf{95.46}($\pm$0.14)\\
\bottomrule
\end{tabular*}
}
\end{minipage}
\end{center}
\vspace{-2.0em}
\end{table}

\begin{table}[H]
\begin{center}
\begin{minipage}{\textwidth}
\caption{Performance comparison with baselines on  training percentages (80\%) 
 (AP)}\label{AP_benchmark}
\begin{tabular*}{\textwidth}{@{\extracolsep{\fill}}lcccccc@{\extracolsep{\fill}}}
\toprule%

Model & Power & HPD &ADV\\
\midrule
Katz~\cite{Katz195339} & 74.29($\pm$0.83)   & 89.52($\pm$0.32)  &93.72($\pm$0.16)  \\
PR~\cite{BRIN1998107}  & 74.74($\pm$0.81)   & 91.01($\pm$0.23)   &94.03($\pm$0.24)\\
SR~\cite{jeh2002simrank} & 70.69($\pm$0.67)   & 84.16($\pm$0.42)    &83.31($\pm$0.35)\\
N2V~\cite{grover2016node2vec}  &76.55($\pm$0.75)   &80.57($\pm$0.81)    &79.02($\pm$0.65)\\
SEAL~\cite{zhang2018link}  & 83.91($\pm$0.83)   & 93.41($\pm$0.09)    &95.18($\pm$0.12)  \\
LGLP~\cite{cai2021line} & 84.78($\pm$0.53)   & 93.65($\pm$0.08)    &95.72($\pm$0.08)\\
LGCL (Ours)  & \textbf{85.46}($\pm$0.62)   & \textbf{94.72}($\pm$0.07)   &\textbf{96.85}($\pm$0.06)\\
\bottomrule
\end{tabular*}
\end{minipage}
\end{center}
\vspace{-3.0em}
\end{table}

\begin{table}[H]
\begin{center}
\begin{minipage}{\textwidth}
\resizebox{\linewidth}{!}{
\begin{tabular*}{\textwidth}{@{\extracolsep{\fill}}lcccccc@{\extracolsep{\fill}}}
\toprule%

Model & SMG & ChCh-Miner & Fdataset \\
\midrule
Katz~\cite{Katz195339} &87.68($\pm$0.90)   & 92.23($\pm$0.21) &93.14($\pm$0.23) \\
PR~\cite{BRIN1998107}   &91.07($\pm$0.59)   & 87.31($\pm$0.12)  &94.68($\pm$0.09)\\
SR~\cite{jeh2002simrank}  &70.39($\pm$1.67)    & 82.54($\pm$0.13)   &87.23($\pm$0.32) \\
N2V~\cite{grover2016node2vec}  &77.01($\pm$1.79)   &78.43($\pm$1.23) &74.42($\pm$0.83)\\
SEAL~\cite{zhang2018link}  &91.90($\pm$0.31)  & 97.46($\pm$0.29)   &92.43($\pm$0.16)\\
LGLP~\cite{cai2021line} &92.92($\pm$0.21)   & 97.33($\pm$0.05) &93.55($\pm$0.14)\\
LGCL (Ours)  &\textbf{93.24}($\pm$0.15)   & \textbf{98.21}($\pm$0.25) &\textbf{94.95}($\pm$0.17)\\
\bottomrule
\end{tabular*}
}
\end{minipage}
\end{center}
\vspace{-2.0em}
\end{table}

\subsubsection{Model Robustness Analysis}
\begin{figure}[H]
\centering
\subfigure{
\includegraphics[width=13.2cm]{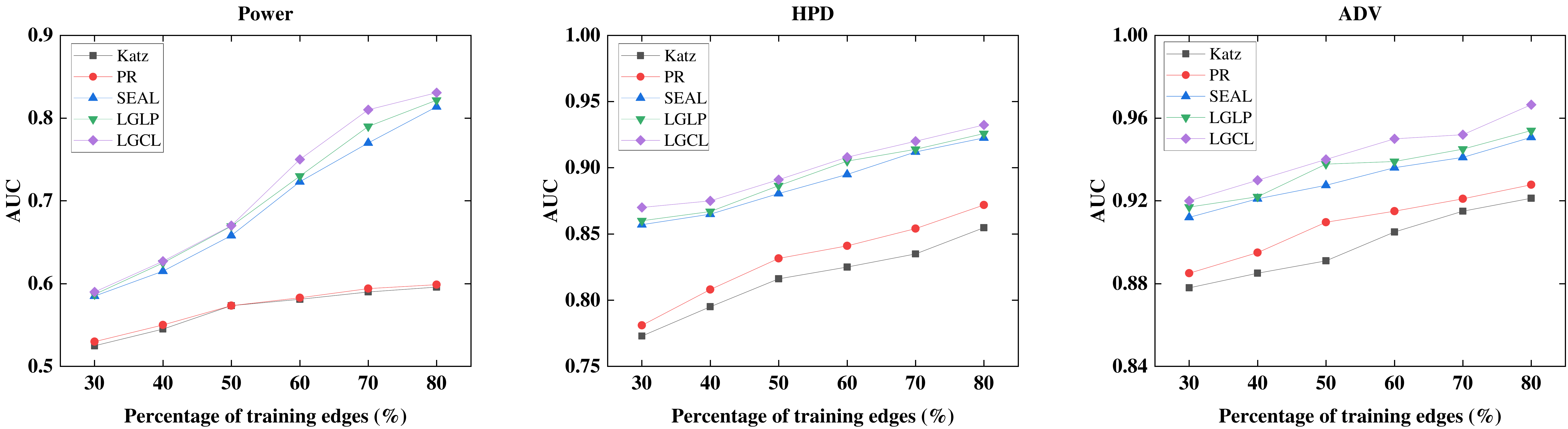}
}
\quad
\subfigure{
\includegraphics[width=13.2cm]{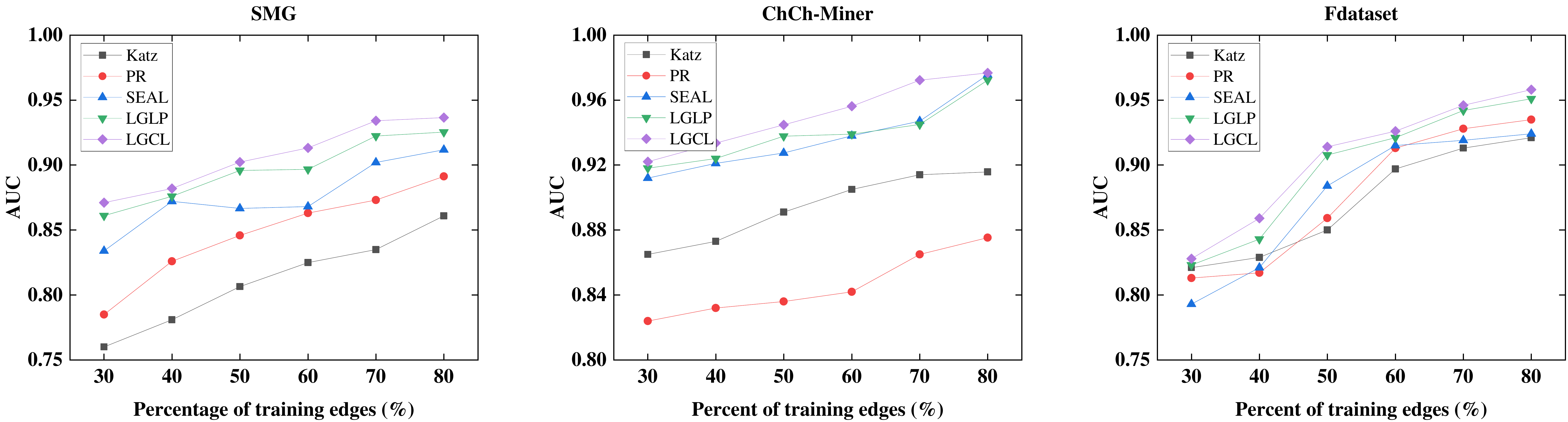}
}
\caption{Robustness analysis on six datasets\label{figRobust}}
\end{figure}

We conduct experiments on edge datasets varying from 30\% to 80\% on six datasets, and the rest of the dataset is used as the test set. The experiments show that LGCL achieves better performance compared with other methods. Fig.\ref{figRobust} shows the robustness of LGCL to network sparsity. To verify the robustness and generalization of the model, we compared it with network similarity methods and graph representation learning methods, and LGCL outperforms them in all assignments with various levels of network sparsity. The performance of Katz and PageRank is poor compared to other methods due to their assumptions. Although LGLP employs line graphs to convert the graph classification task into a node classification task, the performance is partially improved, it relies on the labels of the data and does not take into account the noise caused by the growth of line graph edges, as shown in the figure, the growth of LGLP slows down as the dataset density and the number of edges increases. Because when the number of edges is large, the line graph transformed from the original graph is close to a complete graph, which produces more noise. So LGCL alleviates the model's over-reliance on labels and the noise problem, and provides a significant improvement.

\subsubsection{Ablation Study}
\begin{figure}[H]
\centering
\subfigure{
\includegraphics[width=13.2cm]{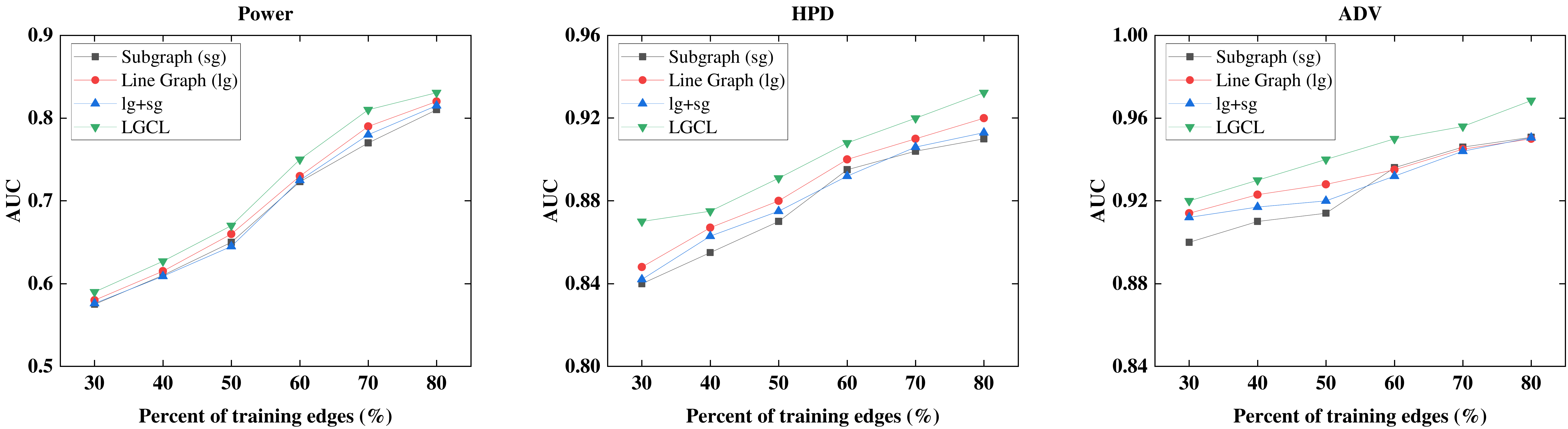}
}
\quad
\subfigure{
\includegraphics[width=13.2cm]{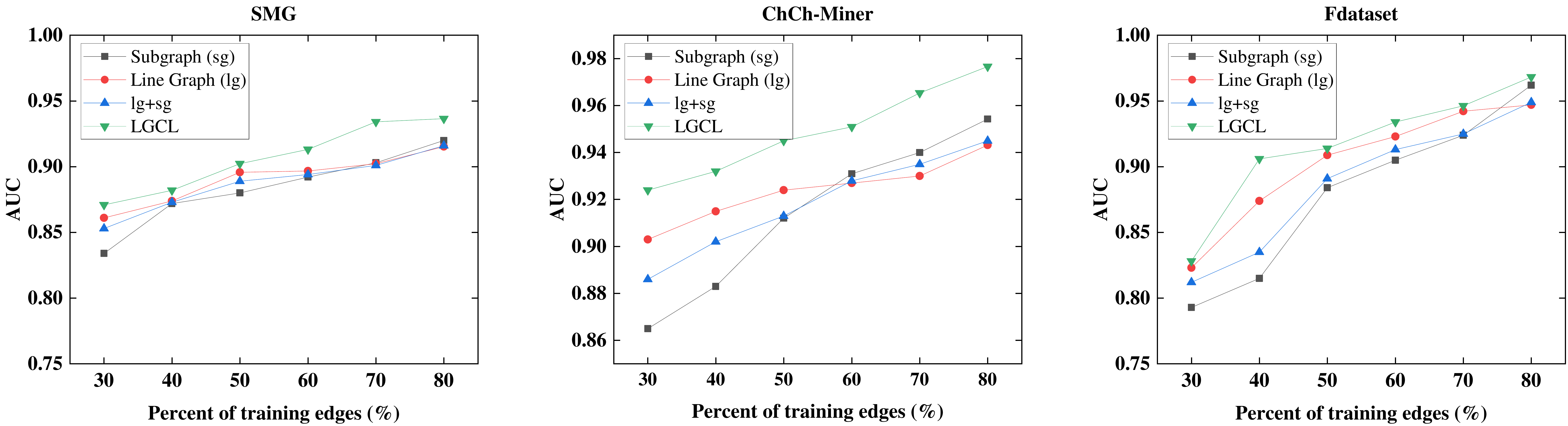}
}
\caption{Results of ablation experiments on six datasets\label{figALB}}
 \vspace{-1.0em}
\end{figure}
To validate the effectiveness of the three components which are subgraph, line graph and contrastive learning by performing experiments on six datasets with different training edge scales from 30\% to 80\%.
As shown in Fig.\ref{figALB}, the result illustrates that the line graph (lg) component which obtains local information performs better than the subgraph (sg) component which obtains global information when the percent of training edges is small, but it does not significantly improve learning capacity as the percentage of train edges increases. Through experiments, we find that the performance of the model can not be significantly improved when the supervised loss of the line graph component is combined with the supervised loss of the subgraph component (lg+sg). By combining the two components with contrastive learning, we can effectively use the information of the two modules to improve the performance of the model. Experiments on ChCh-Miner dataset show that when more than 60\% of the edges are employed, the subgraph (sg) component works better, while the percentage of training sets is less than 60\%, the line graph (lg) component can produce better results. LGCL overcomes the shortcomings of a single view by integrating two views to improve the robustness of the model, and it achieves better performance.

\subsection{Parametric Sensitivity Analysis}
\begin{figure}[H]
 \vspace{-1.0em}
\centering
\includegraphics[width=10cm]{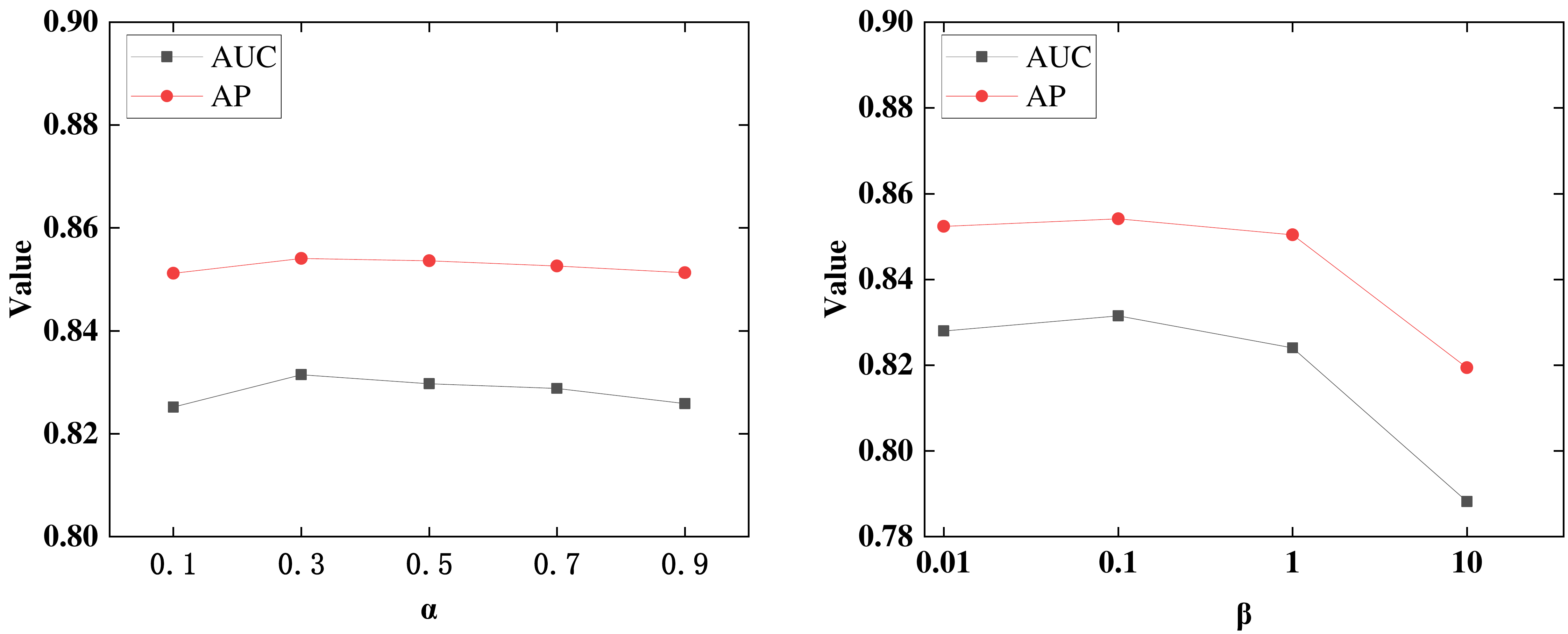}
\caption{Parameter sensitivity study on Power. AUC and AP comparison of LGCL's performance with different parameters.\label{figSA}}
\vspace{-1.0em}
\end{figure}

In our framework in Eq. (\ref{eq_total}), there are two major hyperparameters $\alpha$ and $\beta$. We evaluate their impact on the Power dataset. Fig.\ref{figSA} shows experimental results obtained by only changing one parameter.

\begin{figure}[H]
\centering
\subfigure[Effect of different parameter values]{
\includegraphics[width=0.45\textwidth]{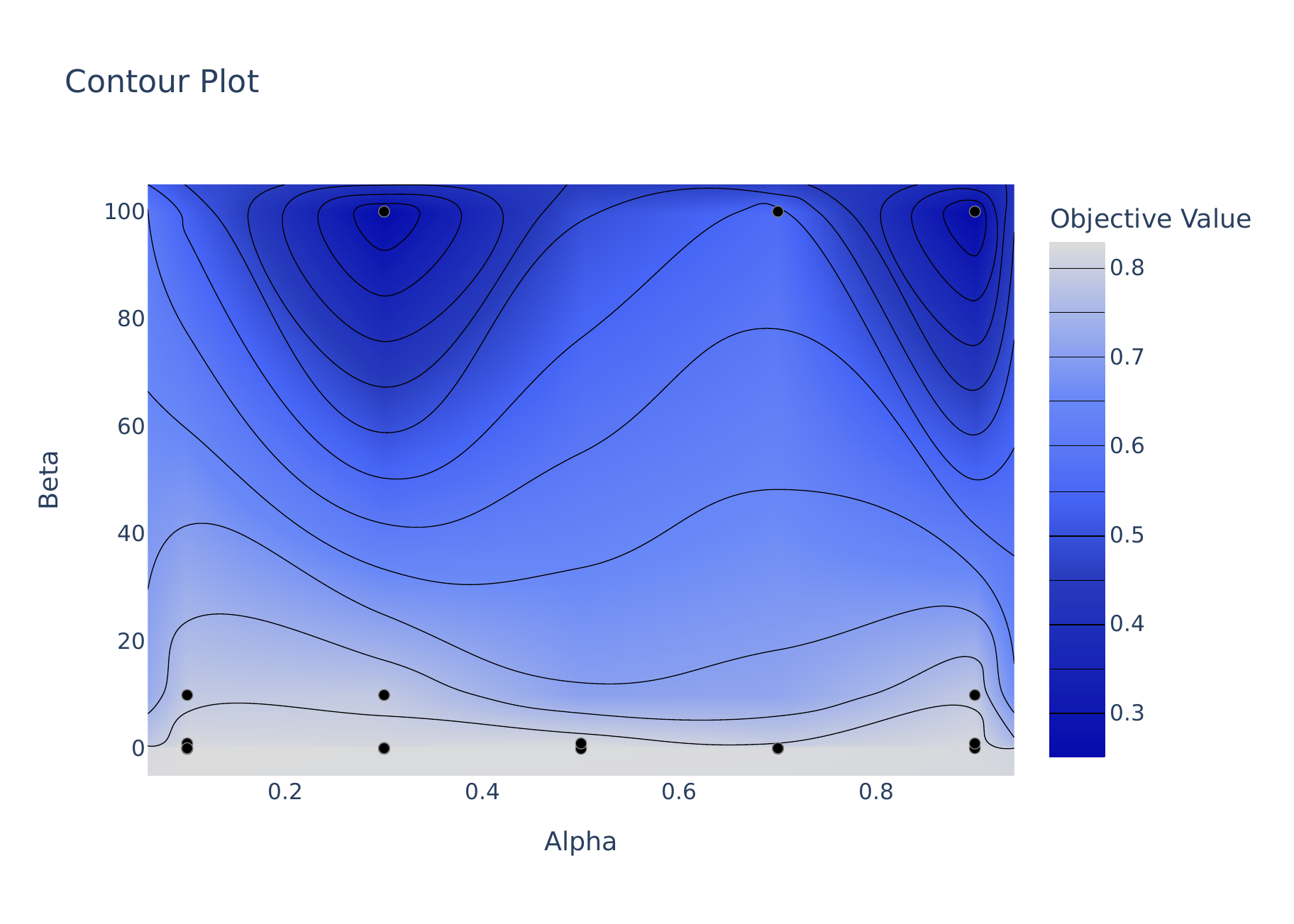}
}
\quad
\subfigure[Parameter importance]{
\includegraphics[width=0.45\textwidth]{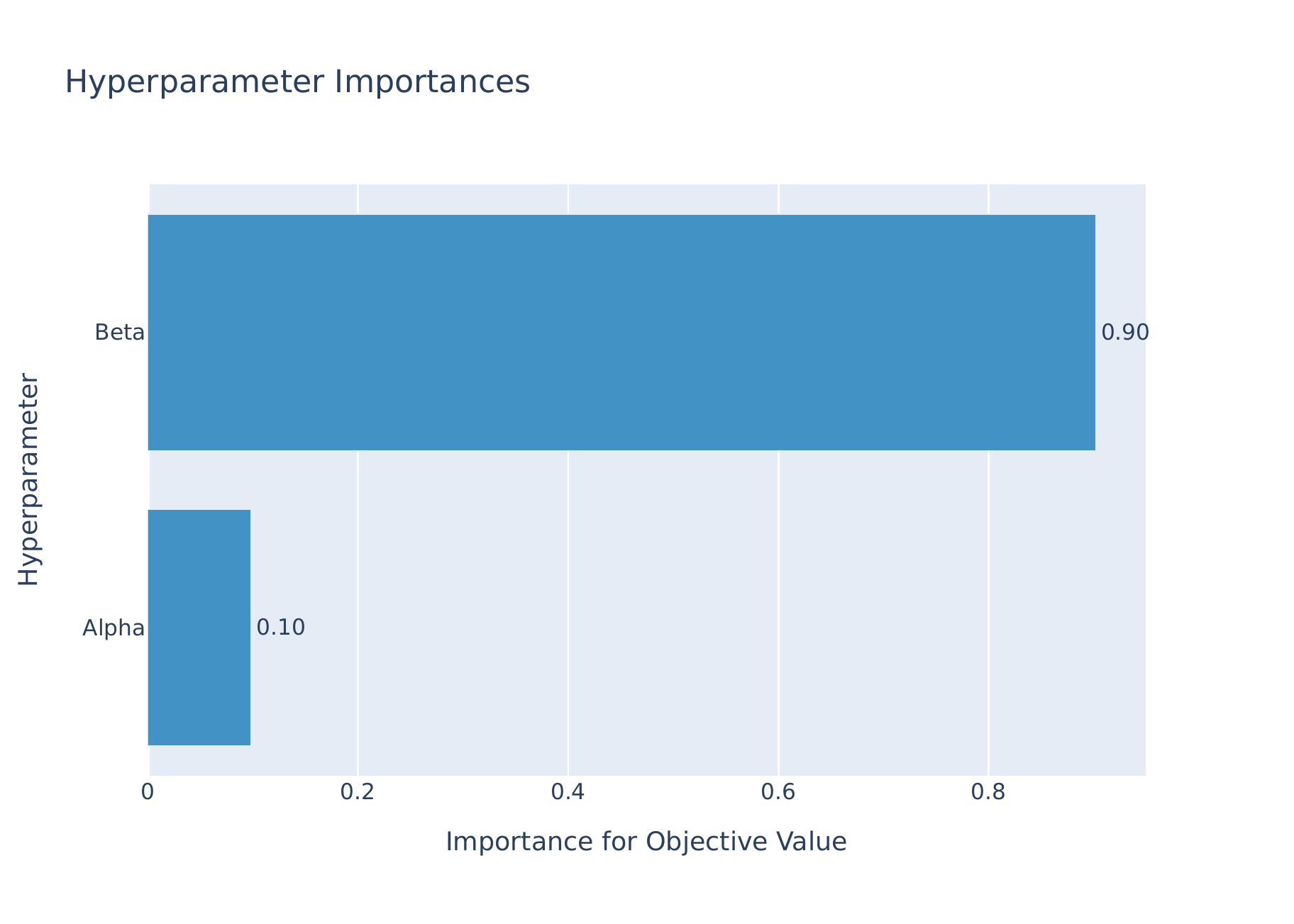}
}
\caption{Optuna uses the TPE (Tree-structured Parzen Estimator) algorithm to sample the parameters and optimize the value of AUC as the objective value, and denotes $\alpha$ and $\beta$ by Alpha and Beta.}
\label{figOp}
\vspace{-1.0em}
\end{figure}

To evaluate the impact of $\alpha$ on model performance, we set $\beta = 0.1$ and vary $\alpha$ from $\{0.1, 0.3, 0.5, 0.7, 0.9\}$. From Fig.\ref{figSA}, we observe that the performance of the the model in AUC and AP is stable over a wide range of $\alpha$, and our model achieves the best performance at $\alpha = 0.3$. Similarly, with $\alpha = 0.3$, we adjust the value of $\beta$ from the range $\{0.01, 0.1, 1, 10\}$. As shown in the figure, the performance of the model is close to optimal performance when $\beta = 0.1$. In summary, the performance of the model is stable for different $\alpha$ and $\beta$.
In order to analyze the relationship between parameters more precisely on the Power dataset, we use Optuna for parameter optimization and analyze the effect of different parameters on AUC. Fig.\ref{figOp} shows that the model works better when $\beta$ is smaller, and the results from Optuna's importance evaluation algorithm, a random forest regression model, indicate that the value of $\beta$ has a relatively large impact on the model performance.
To measure the sensitivity of the LGCL to batch size, we use various batch sizes conduct experiments on HPD and SMG datasets. As shown in the Table~\ref{batch_size}, the performance of our model is less affected by the batch size.
\begin{table}[H]
\vspace{-1.0em}
\begin{center}
\begin{minipage}{\textwidth}
\caption{AUC comparison for LGCL with different batch sizes (70\% training links)}\label{batch_size}
\begin{tabular*}{\textwidth}{@{\extracolsep{\fill}}lcccccc@{\extracolsep{\fill}}}
\toprule%
 & \multicolumn{4}{@{}c@{}}{\quad \quad \quad \quad \quad \quad \quad \quad \quad  \quad \quad batch size }\\\cmidrule{3-6}%
Model & Dataset & 64&128 &256 &512 \\
\midrule
\multirow{2}*{LGCL} &HPD &0.9103	&0.9166	&0.9122	&0.9085\\ &SMG &0.9334	&0.9321	&0.9344	&0.9315 \\
\bottomrule
\end{tabular*}
\end{minipage}
\end{center}
\vspace{-3.0em}
\end{table}

\subsection{Visualization Analysis}
\begin{figure}[H]
     \vspace{-1.0em}
\centering
\subfigure[LGCL]{
\includegraphics[width=3.93cm]{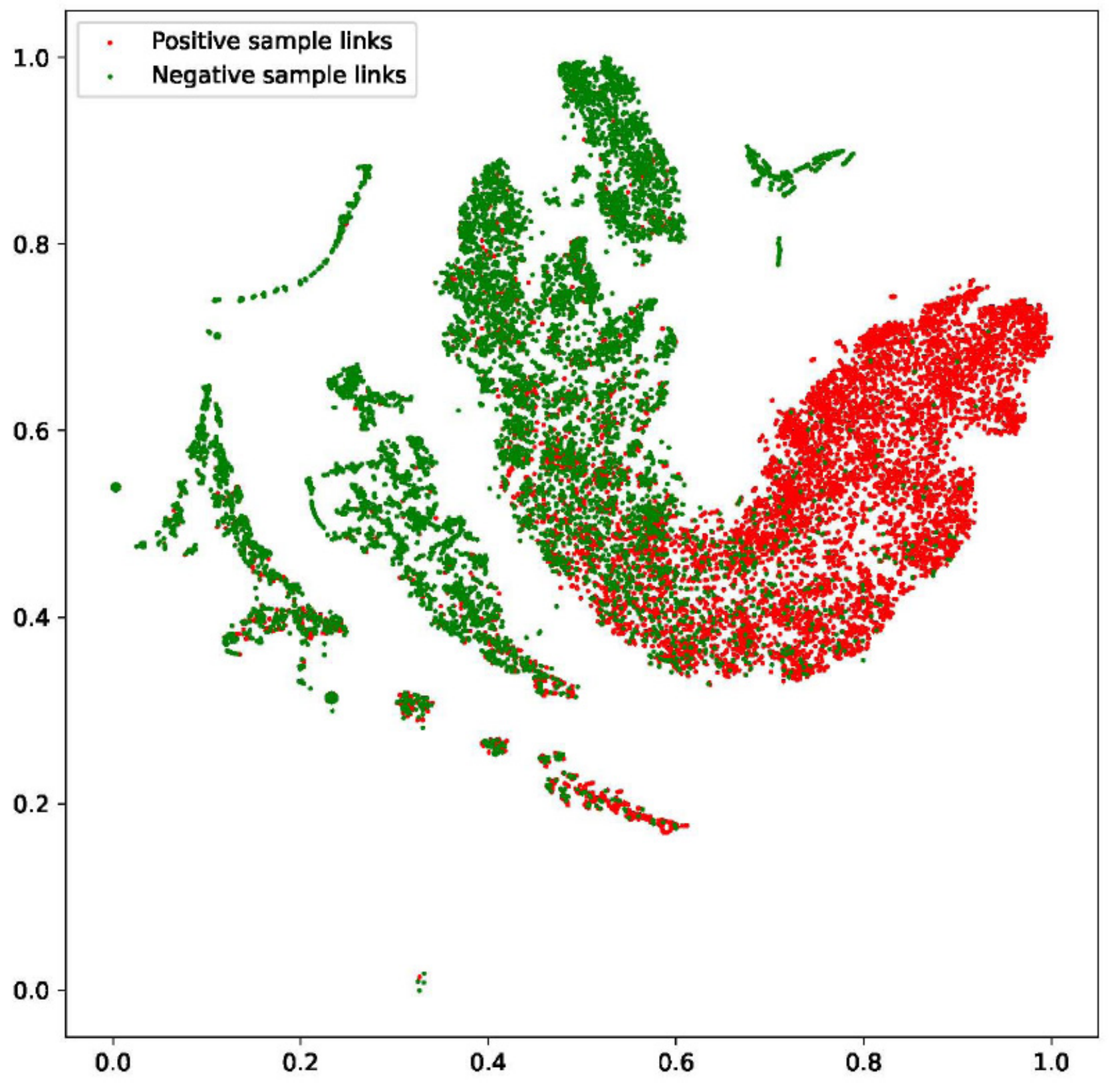}
}
\quad
\subfigure[LGLP]{
\includegraphics[width=3.93cm]{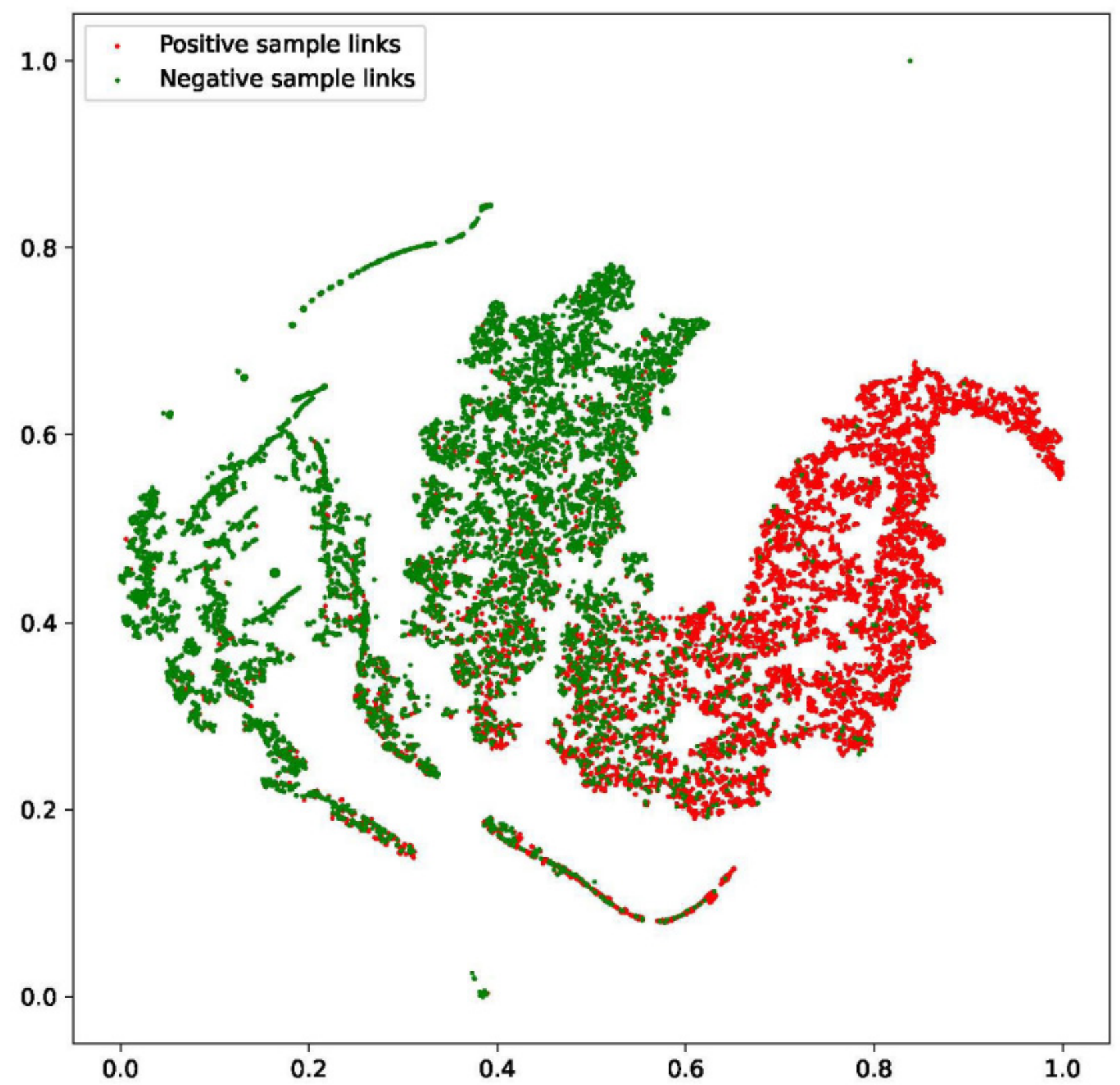}
}
\quad
\subfigure[SEAL]{
\includegraphics[width=3.93cm]{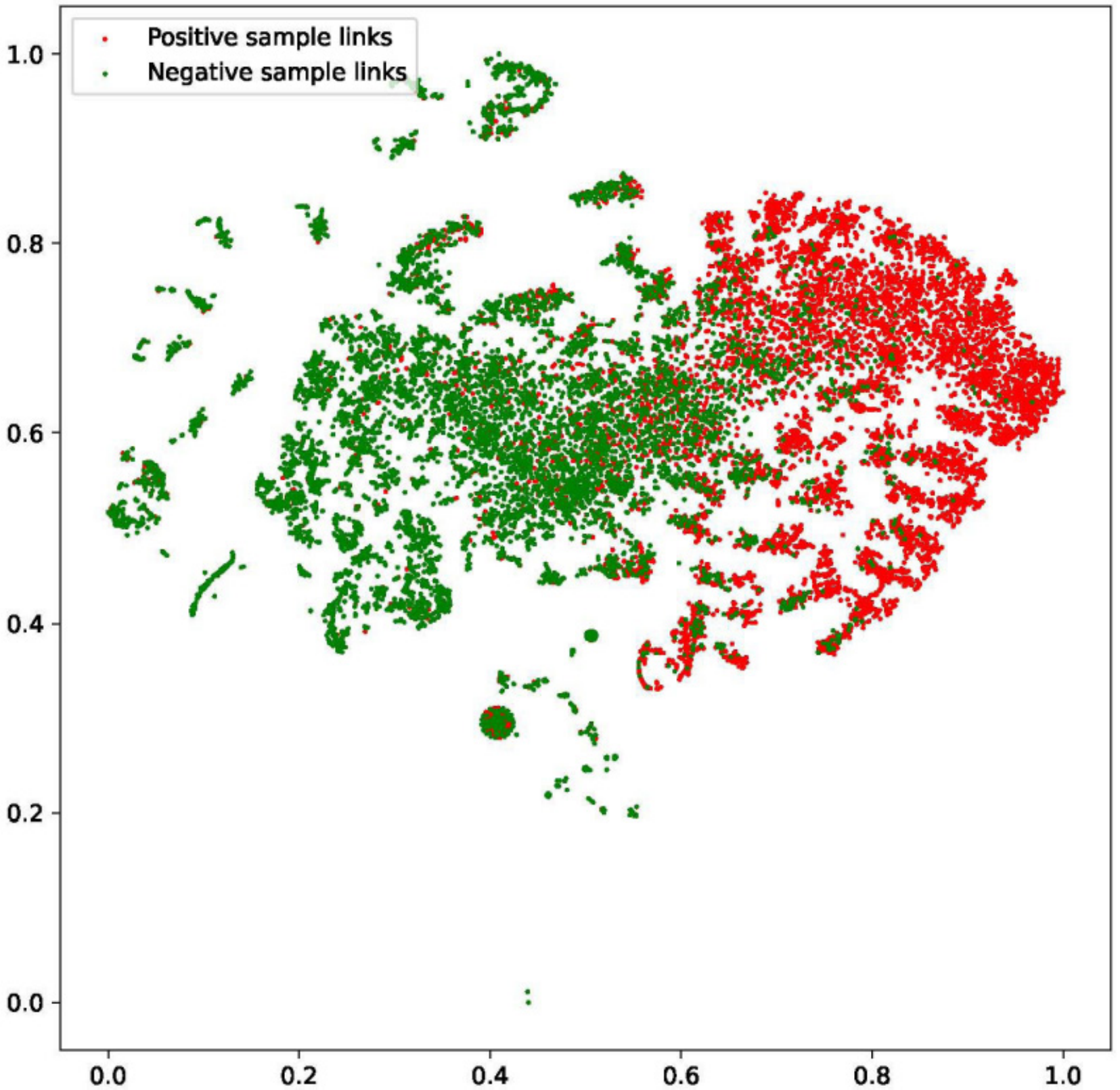}
}\caption{Visualization of HPD dataset, positive sample links are represented by red dots and negative sample links are represented by green dots.}
\label{tsne}
\end{figure}

In this subsection, we use 70\% of the links as the training set and the rest as the test set to perform a visual analysis of the link representation. We extract the link representation learned by the trained model and project it to a 2D space using t-SNE~\cite{van2008visualizing}. We use the HPD dataset as an example to compare LGCL with LGLP, SEAL models. As shown in the Fig.\ref{tsne}, LGLP using only line graph identifies most of the positive sample links, while SEAL using only subgraphs has poor differentiation effect. LGCL balances the information of line graph nodes and subgraphs by contrastive learning, and realizes cross-scale contrast, it has good classification effect on two kinds of links, positive sample links and negative sample links, which makes the two kinds of links easier to distinguish.
\section{Conclusion}

In this paper, we propose a  contrastive learning method to balance the information of line graphs and extracted subgraphs.  After converting the graph to a line graph, the link prediction task is transfered into a node classification task, which can directly take advantage of graph convolution operator on node embedding learning. The proposed LGCL method as the natural cross-scale learning progress can contrast subgraphs with line graph nodes. And the information on different perspectives is taken into account, which enhances the robustness of the model.

Besides, there are still many problems that we need to continue to explore in future works, such as the information difference between different graph structures after being converted into line graphs, the elimination of redundant edges of line graphs, and the optimization of time complexity of line graphs in graph neural network information transmission.
\section*{Acknowledgements}
This work was supported by the National Natural Science Foundation
of China (62172242, 51901152), Industry University Cooperation Education
Program of the Ministry of Education (2020021680113) and Shanxi Scholarship Council of China.




\bibliographystyle{elsarticle-num} 
\newpage
 \textbf{Zehua Zhang} received the Ph.D. degree
 from Tongji University, Shanghai, China in
 2014. He is currently an associate professor
 at Taiyuan University of Technology. His research interests include bio-feature recognition and applications, social networks, complex network pattern analysis.\par

  \textbf{Shilin Sun}  received the B.S. degree from Xinjiang University, Urumqi, China in 2019. He is currently a M.S. candidate at Taiyuan University of Technology from 2020. His research is focused on the application of graph embeddings for biomedical data mining.\par
    \textbf{Guixiang Ma} received the PhD degree in
Computer Science from University of Illinois at
Chicago in 2019. She is currently an AI Research
Scientist at Intel Labs. Her research interests
include machine learning, data mining, graph
representation learning and their applications in
various domains.\par
  \textbf{Caiming Zhong} is a professor in College of Science and Technology, Ningbo University, Ningbo, China. His research interests include cluster analysis, manifold learning and image segmentation.\par
\end{document}